\pdfoutput=1
\documentclass[11pt]{article}

\usepackage[]{acl}

\usepackage{times}
\usepackage{latexsym}

\usepackage[T1]{fontenc}
\usepackage[utf8]{inputenc}

\usepackage{microtype}

\usepackage{algorithm,algpseudocode}
\usepackage{amsmath}
\usepackage{amssymb}
\usepackage{graphicx}
\usepackage{subcaption}
\usepackage{multicol}
\usepackage{url}
\usepackage{bbm}
\usepackage{stfloats}
\usepackage{flafter}

\title{Uncertainty Sentence Sampling by Virtual Adversarial Perturbation}

\author{Hanshan Zhang  \and Zhen Zhang  \and Hongfei Jiang \and Yang Song \\
  Zuoyebang Education Technology (Beijing) Co., Ltd  \\
  \texttt{\{zhanghanshan,zhangzhen,jianghongfei,songyang\}@zuoyebang.com}}

\begin{document}
\maketitle
\begin{abstract}
Active learning for sentence understanding attempts to reduce the annotation cost by identifying the most informative examples. Common methods for active learning use either uncertainty or diversity sampling in the pool-based scenario. In this work, to incorporate both predictive uncertainty and sample diversity, we propose Virtual Adversarial Perturbation for Active Learning (VAPAL) , an uncertainty-diversity combination framework, using virtual adversarial perturbation \cite{Miyato2019} as model uncertainty representation. VAPAL consistently performs equally well or even better than the strong baselines on four sentence understanding datasets: AGNEWS, IMDB, PUBMED, and SST-2, offering a potential option for active learning on sentence understanding tasks.
\end{abstract}

\section{Introduction}
%重写 扔nlp角度 
In recent years, deep neural networks have made a significant achievement in natural language processing \cite{Yang2019,Devlin2019,Raffel2020,He2020}. These neural models usually need a large amount of data which have to be labeled and trained. Active learning is an effective way to reduce both computational costs and human labor by selecting the most critical examples to label.

Two samplings, uncertainty sampling and diversity sampling are often used in common active learning methods. Uncertainty sampling \cite{Lewis1994} selects difficult examples based on the model confidence score. In a batch setting, the sampled data points are near-identical \cite{Ash2019DeepBounds}. This phenomenon suggests that we might need to take diversity into account besides uncertainty. A naive combination of uncertainty and diversity, however, negatively affects the test accuracy \cite{Hsu2015}. \citet{Ash2019DeepBounds} presents a practical framework, BADGE, which combines uncertainty and diversity. BADGE measures uncertainty by gradient embedding and achieves diversity by clustering.

However, BADGE relies on model confidence scores, which require a model warm-start to calibrate. Specifically, the correctness likelihoods do not increase consistently with higher model confidence scores. To avoid the warm-start requirement,  \citet{Yuan2020Cold-startModeling} presents ALPS, a cold-start approach that
uses self-supervised loss (Masked Language Model Loss) as sentence representation. Nevertheless, MLM loss can be seen as a language model perplexity, not a direct downstream task-related measurement.

From another point of view, deep learning models are vulnerable to adversarial examples \cite{Goodfellow2014,Kurakin2016}, indicating that measuring uncertainty based on model confidence scores is overconfident. Virtual adversarial training \cite{Miyato2015,Miyato2019,Miyato2016} modifies inputs with special perturbations, virtual adversarial perturbation, which can change the output distribution of the model in the most significant way in the sense of KL-divergence. It is tested valid in industry-scale semi-supervised learning setting \cite{Chen2021IndustryUnderstanding}. 

We propose VAPAL (Virtual Adversarial Perturbation for Active Learning) in this work. VAPAL computes perturbation for each data point in an unlabeled pool to measure model uncertainty. VAPAL clusters the data points to achieve diversity like BADGE and ALPS with perturbations acquired. Since virtual adversarial perturbation could be calculated without label information, our method VAPAL is also advantageous over BADGE in that it does not require a warm-start. Unlike ALPS, our method does not rely on a special self-supervised loss. In other words, VAPAL could be applied to any differentiable model. 

We use four datasets (AGNEWS, IMDB, PUBMED, and SST-2) to evaluate VAPAL through two tasks, namely sentiment analysis and topic classification. Baselines cover uncertainty, diversity, and two SOTA hybrid active learning methods (BADGE, ALPS). 

Our main contributions are as follows:
\begin{itemize}
    \item We take Virtual Adversarial Perturbation to measure model uncertainty in sentence understanding tasks for the first time. The local smoothness is treated as model uncertainty, which relies less on the poorly calibrated model confidence scores.
    \item We present VAPAL (Virtual Adversarial Perturbation for Active Learning) to combine uncertainty and diversity in a combination framework.
    \item We show that VAPAL method is equally good or better than the baselines in four tasks. Our methods can successfully replace the gradient-based representation of BADGE. Furthermore, it does not rely on specific self-supervised loss, unlike Masked Language Model Loss used in ALPS.
\end{itemize}

\section{Related Work}
To reduce the cost of labeling, active learning seeks to select the most informative data points from the unlabeled data to require humans to obtain labels. We then train the learner model on the new labeled data and repeats. The prior active learning sampling methods primarily focus on uncertainty or diversity. The uncertainty sampling methods are the most popular and widely used strategies which select the difficult examples to label \cite{Lewis1994,5206627,Houlsby2011}. Diversity sampling selects a subset of data points from the pool to effectively represent the whole pool distribution \cite{Geifman2017DeepTail,Sener2017ActiveApproach,Gissin2019}. A successful active learning method requires the incorporation of both aspects, but the exact implementation is still open for discussion.
% the exact implementation of integration of both aspects necessary for a successful active learning method is still an open discussion.

Recently, hybrid approaches that combine uncertainty and diversity sampling have also been proposed. Naive combination frameworks are shown to be harmful to the test accuracy and rely on hyperparameters \cite{Hsu2015}. Aiming for sophisticated combination frameworks, \citet{Ash2019DeepBounds} propose Batch Active Learning By Diverse Gradient Embeddings (BADGE), and \citet{Yuan2020Cold-startModeling} propose Active Learning by Processing Surprisal (ALPS).  They follow the same framework that first builds uncertainty representations for unlabeled data and clustering for diversity. 
% BADGE builds data uncertainty representations by gradient embeddings formed by the model last layer parameters gradients respecting the cross-entropy loss.
BADGE measures data uncertainty as the gradient magnitude with respect to parameters in the final (output) layer and forms a gradient embedding based data representation. However, according to \citet{Yuan2020Cold-startModeling}, BADGE has two main issues: reliance on warm-starting and computational inefficiency. ALPS builds data embeddings from self-supervised loss (Masked Language Model loss) \cite{Yuan2020Cold-startModeling}.  Nevertheless, the MLM loss is an indirect proxy for model uncertainty in the downstream classification tasks, and ALPS might work only with a pre-trained language model using MLM. 

What else can be used as a model uncertainty representation and can be efficiently combined to achieve diversity sampling? Virtual adversarial perturbation from virtual adversarial training \cite{Miyato2019} is a promising option. Deep learning methods often face possible over-fitting in model generalization, especially when the training set is relatively small. In adversarial training, adversarial attacks are utilized to approximate the smallest perturbation for a given latent state to cross the decision boundary \cite{Goodfellow2014,Kurakin2016}. It has proven to be an important proxy for assessing the robustness of the model. Moreover, the labeled data is scarce. Virtual adversarial training (VAT) does not require true label information, thus fully using the unlabeled data. VAT can be seen as a regularization method based on Local Distributional Smoothness (LDS) loss. LDS is defined as a negative measure of the smoothness of the model distribution facing local perturbation around input data points in the sense of KL-divergence \cite{Miyato2019}. The virtual adversarial perturbation can be crafted without the label information, which can help alleviate the warm-starting issue BADGE faces. \citet{Yu2020VirtualLearning} roughly rank grouped examples of model prediction by LDS score. Our method is inspired by the same research line \cite{Miyato2019} but different from it in many folds. Our method aims to project data into a model smoothness representation space rather than a rough scalar score, so it is more effective. We introduce the virtual adversarial perturbation as sentence representations by which model uncertainty is inherently expressed. Furthermore, we consider both uncertainty and diversity in a uniform framework based on this informative representation. 

\section{Methods}

\subsection{Notation}

The notation for the sake of active learning tasks will be introduced in the following section. First, each sentence $x_i$ is a sequence of tokens: 
\begin{small}
\begin{equation}
x_i = <t_0,\dots,t_s>, t_s \in \mathcal{V}
\end{equation}
\end{small}

where $t_s$ is tokens in sentence, $\mathcal{V}$ is vocabulary  set and $x_i$ is from a discrete space $\mathcal{D}$. Second, for labeled examples, 
\begin{small}
\begin{equation}
\mathcal{L} = \{(x_{i},y_{i}) \mid x_i \in \mathcal{D}, y_i \in \mathcal{Y}, i = \{1 ,\dots , L\} \} 
\end{equation}
\end{small}
Where $\mathcal{L}$ is labeled set for classification task and the labels belongs to set $\mathcal{Y} = \{1, \dots ,C\}$. Except labeled set, we also have an unlabeled data pool,namely,
\begin{small}
\begin{equation}
\mathcal{U} = \{(x_{u}) \mid x_u \in \mathcal{D},i=\{1 ,\dots , U\} \} 
\end{equation}
\end{small}
Thus, for our total data set, $\mathcal{D} = \mathcal{L}+\mathcal{U}$.

Let $f(x;\theta)=\sigma(V \cdot h(x;W)):\mathcal{D} \to \mathcal{Y}$ with parameters $\theta = (W,V)$. We fine-tune learner model $f(x;\theta)$ on the labeled set $\mathcal{L}$ with pre-trained model BERT, $h(x;W)$ and the attached sequences classification head ($V$).

\subsection{Virtual Adversarial Perturbation}

Using model smoothness as a regularization strategy to avoid over-fitting during model training is an effective way in general. Vitual Adversarial perturbation is introduced by \citet{Miyato2015,Miyato2019,Miyato2016} for calculating the local smoothness of the conditional label distribution around each input data point. With small perturbation \textbf{r}, we can compute the KL-divergence as follows:

\begin{equation}
    D_{KL}(\textbf{r},x_i,\theta) = D_{KL}(p(y \mid x_i, \theta) || p(y \mid x_i + \textbf{r},\theta))
\end{equation}
\begin{equation}
    \textbf{r}_{v-adv}^i = \operatorname{argmax}\{D_{KL}(\textbf{r},x_i,\theta) : \|\textbf{r} \|_2 \leq \epsilon \}
\end{equation}

where $\mathbf{r}_{v-adv}^i$ is the optimal virtual adversarial perturbation for data point $x_i$. Note $\mathbf{r}_{v-adv}^i$ is the most sensitive direction of model distribution $p(y|x_i,\theta)$ in the sense of KL-divergence. As for finding optimal $\mathbf{r}_{v-adv}$, we can approximate $\mathbf{r}_{v-adv}$ with the power iteration method \cite{GOLUB200035}:

\begin{equation}
d \gets \overline{\Delta_r D_{KL}(\textbf{r}_{v-adv}^i,x_i,\hat{\theta})\mid_{r=\xi d}}
\end{equation}

With a larger number of power iteration, $I_p$, the approximation can be improved monotonically.

LDS (Local Distributional Smoothness) loss can be calculated \cite{Miyato2015,Miyato2019,Miyato2016} as :
\begin{equation}
LDS(x_i,\theta) = -D_{KL}(\mathbf{r}_{v-adv}^i,x_i,\theta)
\end{equation}

The loss can be thought as a negative measure of model local smoothness given input $x_i$.

\subsection{Virtual Adversarial Perturbation for Activate (VAPAL) }

\begin{algorithm}[ht]
  \caption{VAPAL: Virtual Adversarial Perturbation for Activate Learning}\label{alg:algorithm1}
  \begin{algorithmic}[1]
    \Require
      \Statex{Neural network $f\left(x;\theta\right)$, unlabeled data pool $U$, number of query size $m$, number of iterations $T$, number of iterations of the power method $I_p$, number of batch size $B$}
    \Statex
    \For{$t \gets 1,\dots,T$}
        
      \For{all examples $x$ in $U \setminus \mathcal{L}_{t-1}$}
      \State 
        Compute $r_{v-adv}^x$ \Comment{ using EQ.\ref{eq:radv1} and \ref{eq:radv2}}
      \EndFor
      
      \State $V_t = \{r_{v-adv}^x \mid x \in U\setminus \mathcal{L}_{t-1} \}$
      
      \State $M \gets \text{k-means} (V_t)_{k=m} $ 
      
        \State $Q=\{\operatorname*{argmin}_{x\in {U \setminus \mathcal{L}_{t-1}}} \|c-r_{v-adv}^x\| \mid c \in M\} $

    \State $\mathcal{L}_t \gets \mathcal{L}_{t-1} \cup Q$
    
    \State Train a model $\theta_{t}$ on $\mathcal{L}_t$
    
    \EndFor
    \State \textbf{return} Final Model $\theta_{T}$
  \end{algorithmic}
\end{algorithm}

VAPAL, described in Algorithm \ref{alg:algorithm1}, starts by drawing an initial set of $m$ examples using K-MEANS on $r_{v-adv}^i$ to find $m$ nearest examples respect to m centers. Its main process has three computations at each step $t$: a $r_{v-adv}$ computation, a sampling computation, and a fine-tune $f(x;\theta)$ computation. In detail, at each time step $t$, for every $x_i \in \mathcal{U}$, we compute the $r_{v-adv}^i$. Given these virtual adversarial perturbations $\{r_{v-adv}^i: x_i \in \mathcal{U}\}$, VAPAL selects a $m$ size set of data points by sampling via $k$-MEANS \cite{Lloyd1982}. The algorithm use these sampled example set asking oracle to get the corresponding labels, retrain the model and repeats.

The more detail of the main computations: a $r_{v-adv}$ computation, a sampling computation, a fine-tune $f(x;\theta)$ computation is discussed in the following sections.

\subsubsection{Modified $r_{v-adv}$ Computation}

Unlike computer vision and speech tasks, it is not-trivial for VAPAL to compute $r_{v-adv}$ directly on sentences(input space) because the text is drawn from a discrete space. Thus we modified the computation as follows:
\begin{equation}
\label{eq:radv1}
D_{KL}(\mathbf{r},h_i,\theta) = D_{KL}(p(y \mid h_i, \theta) || p(y \mid h_i + \mathbf{r},\theta))
\end{equation}

\begin{equation}
\label{eq:radv2}
d \gets \overline{\Delta_r D_{KL}(\mathbf{r}_{v-adv}^i,h_i,\hat{\theta})\mid_{r=\xi d}}
\end{equation}

where $h_i = h(x_i;W)$. Instead of applying attack on input discrete space, we use the $[CLS]$ token embedding from pre-trained BERT encoder $h(x_i;W)$ to represent data points in $\mathcal{D}$. So the  $r_{v-adv} $ is in $\mathbb{R}^d$ space. The $r_{v-adv}^i$ is thought as a measurement of model uncertain about example $x_i$ in the sense of KL-divergence which don't need to consider $\hat{y}$ as the true label. Thus the first limitation of BADGE: reliance on warm-starting is not taking a major impact on VAPAL.  Considering the second limitation of BADGE, computational inefficiency, the gradient embeddings in BADGE is $g_x \in \mathbb{R}^{Cd}$ where $C$ is the number of labels and $d$ input hidden dimension of the last project layer. In contrast, $r_{v-adv}$ is only $d$ dimension, which largely reduces the cost of computing distances of clustering step.

\subsubsection{The Sampling step: $k$-MEANS Clustering}

After computing the virtual adversarial perturbations set $V_t=\{r_{v-adv}^x \mid x \in \mathcal{U}\}$ in the unlabeled pool, we apply $k$-MEANS to cluster the perturbations (from our experiment, $k$-MEANS is  slightly better than $k$-MEANS++).  We select $m$ data points whose virtual adversarial perturbation is nearest to $m$ cluster centers. The final set of sentences $Q$ are the queries for oracle to label. 

\subsubsection{Fine-Tuning}

We sample a batch of $m=100$ examples from the training data set, query the labels, and then move the labeled set out of the unlabeled data pool for active learning simulation. The initial BERT encoder $h(x;W)$ is a pre-trained BERT-based model. We fine-tune the classifier $f(x;\theta_t)$ with labeled data set and report accuracy performance on the test set for each iteration. To avoid warm-starting \cite{Ash2020}, we don't continue fine-tune the model $f(x;\theta_{t-1})$ from previous iteration. The total iteration is set to be 10. A total of 10000 examples is collected for each classification task in the end. 

\section{Experiments}
\label{Experiments}
We evaluate VAPAL on sentiment analysis, sentence classification for three different domains:  news articles, medical abstracts,  and sentiment reviews. For the labeling, we directly use the label information in four data sets. In all experiments, we report the averaged test F1 score result of five runs with different random seeds.

\subsection{Dataset}

We follow \cite{Yuan2020Cold-startModeling} and use IMDB \cite{maas-etal-2011-learning},SST-2 \cite{Socher2013ReasoningCompletion}, PUBMED \cite{Dernoncourt2017} and AGNEWS \cite{Zhang2015}.  We summarised four data sets' statistics in Table.\ref{table:Summary of the Sentence Classification datasets}.  
\begin{table*}[t]
\begin{center}
\resizebox{.5\textwidth}{!}{%
\begin{tabular}{ |c|c|c|c|c|c|c|c| } 
   \hline
  \textbf{Dataset} & \textbf{Domain} & \textbf{\#Train} & \textbf{\#Dev} & \textbf{\#Test} & \textbf{\#Label} \\
   \hline
   AG NEWS   & News Articles &110k &10k & 7.6k & 4 \\
    IMDB &  Sentiment Reviews & 17.5k &7.5k &25k & 2 \\
    PUBMED 20K RCT & Medical Abstracts & 180k & 32.2k & 30.1k & 5 \\
    SST-2 & Sentiment Reviews & 60.6k & 6.7k & 873 &2 \\
    \hline
\end{tabular}%
}
\end{center}
  \caption{Summary of used Sentence Classification data sets}
  \label{table:Summary of the Sentence Classification datasets}
\end{table*}

\subsection{Baselines}

We compare VAPAL against four baseline methods: three warm-start methods (Entropy, BADGE), and two cold-start methods (RAND,ALPS):

\textbf{RAND} directly sample data points from the unlabeled pool, $\mathcal{U}$, with a uniform distribution.

\textbf{Entropy} select data examples with highest predictive entropy with model \cite{Lewis1994,Wang2014}

\textbf{BADGE} \cite{Ash2019DeepBounds}, computes loss gradient embeddings $g_x=\frac{\partial}{\partial \theta_{out}}\mathcal{L}_{CE}(f(x; \theta), \hat{y}_x)$ where $\hat{y}_x=\operatorname*{argmax}_{y\in \mathcal{Y}} f(x;\theta_t)_y$. Then $g_{x_i}$ is
\begin{small}
\begin{equation}
(g_{x_i})= (f(x:\theta) - \mathbbm{1}(\hat{y}=i))h(x;W)
\end{equation}
\end{small}
for each data point in unlabeled data pool $\mathcal{U}$ respecting to the model's last layer parameters. Then select batch set of data by using cluster centers  of $k$-MEANS++ \cite{Arthur2007K-Means++:Seeding} on the gradient embedding set for labeling. 

\textbf{ALPS} \cite{Yuan2020Cold-startModeling}  passes the unmasked sequence example $x_i$ through BERT and MLM head, randomly samples 15\% tokens to compute the cross-entropy loss against the target labels. Using these MLM losses for each data point in the pool, they use $k$-MEANS to select a batch set of examples for labeling.

Data is randomly sampled at start for the warm-start method. The implement of these four methods is based on the code from \cite{Yuan2020Cold-startModeling} \footnote{\url{https://github.com/forest-snow/alps}}

\subsection{Implementation Details}

The batch size for fine-tuning is set to be 32 with three epochs. The max sequence length is set to be 128 for all data sets. We use AdamW \cite{Loshchilov2019} with learning rate of $2e-5$, $\beta_1=0.9$,$\beta_2=0.99$ and no warm-up learning rate schedule. For calculating $r_{v-adv}$, the power of iteration $I_p$ is 10,  $\epsilon=1$ and $\xi=10$. 

Following \cite{Yuan2020Cold-startModeling}, the uncased BERT-BASE\footnote{\url{https://huggingface.co/transformers/}} is used as encoder $h(x;\theta)$ for three tasks, IMDB,SST-2,AGNEWS. For PUBMED, the pretrained BERT is from SCIBERT \cite{Beltagy2019}\footnote{\url{https://github.com/allenai/scibert}}. All experiments are run on one GeForce GTX 2080 GPU and 2.50GHz Intel(R) Xeon(R) Platinum 8255C CPU.

\section{Result and Discussion}
\label{sec:Result}
\begin{figure*}[ht]
\begin{subfigure}[b]{.5\textwidth}
  \centering
  % include first image
  \includegraphics[width=\linewidth]{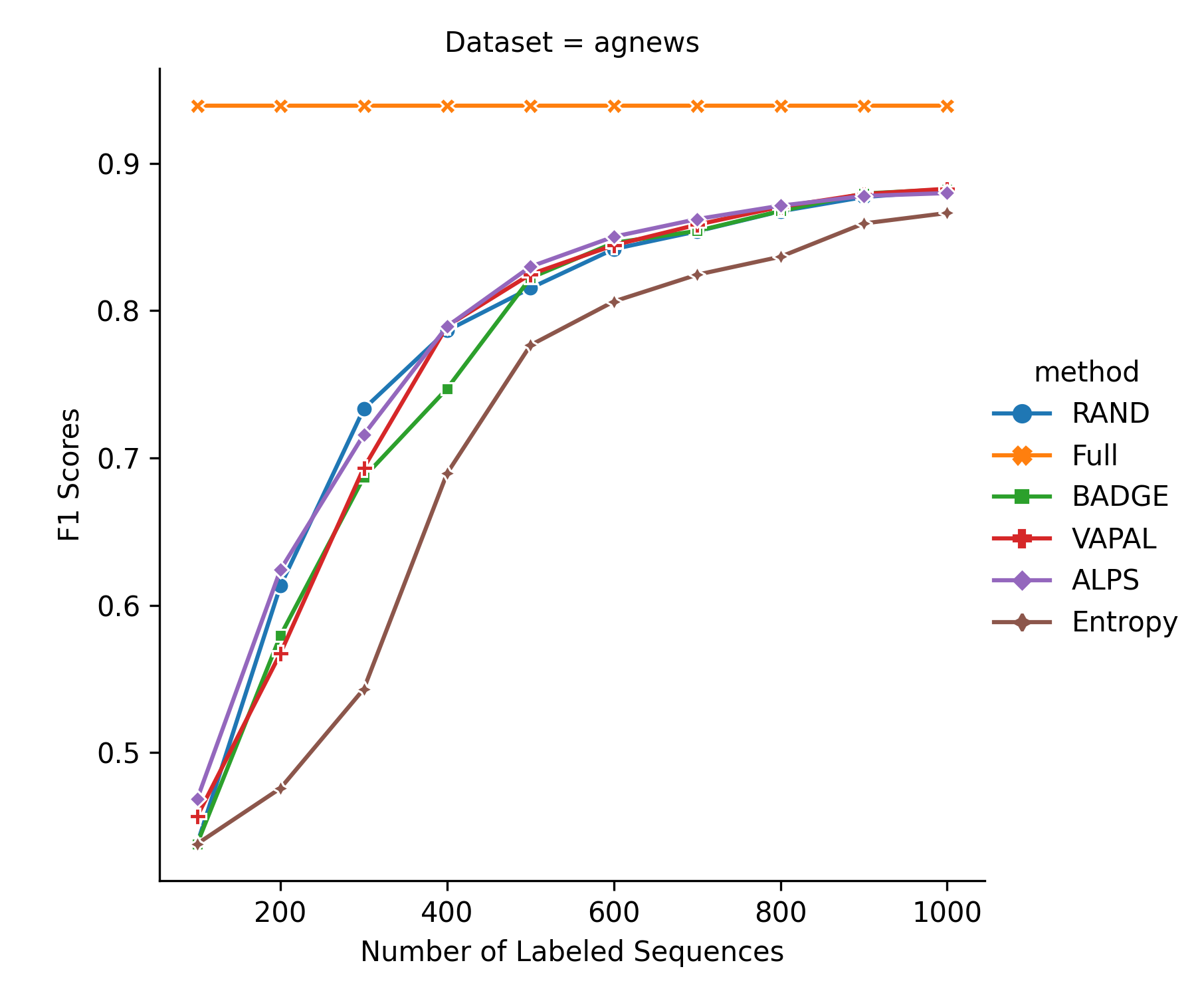}  
  \caption{Dataset: AGNEWS}
  \label{fig:whole-agnews}
\end{subfigure}%
\begin{subfigure}[b]{.5\textwidth}
  \centering
  % include second image
  \includegraphics[width=\linewidth]{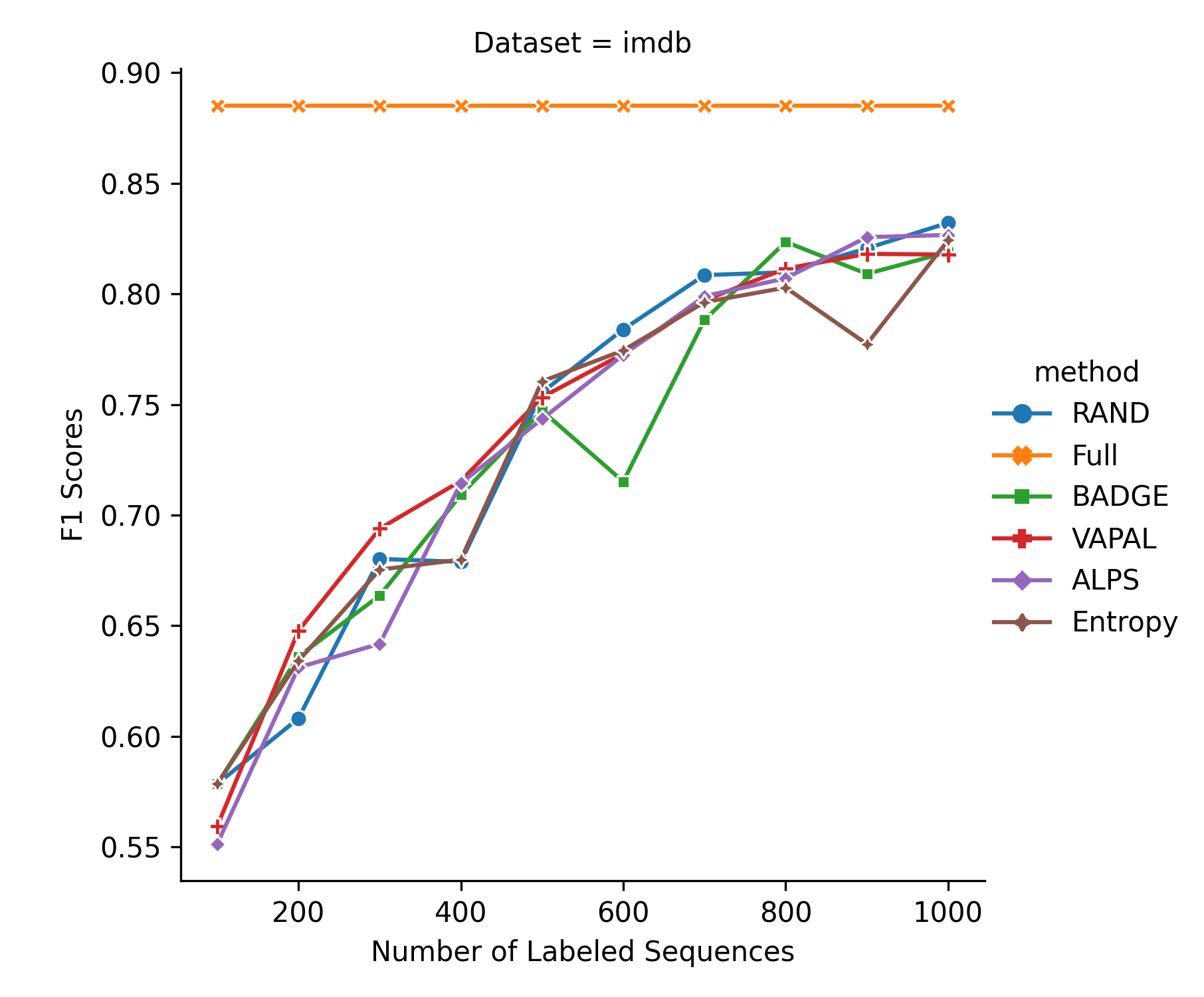}
  \caption{Dataset: IMDB}
  \label{fig:whole-imdb}
\end{subfigure}%
\newline
\begin{subfigure}[b]{.5\textwidth}
  \centering
  % include third image
  \includegraphics[width=\linewidth]{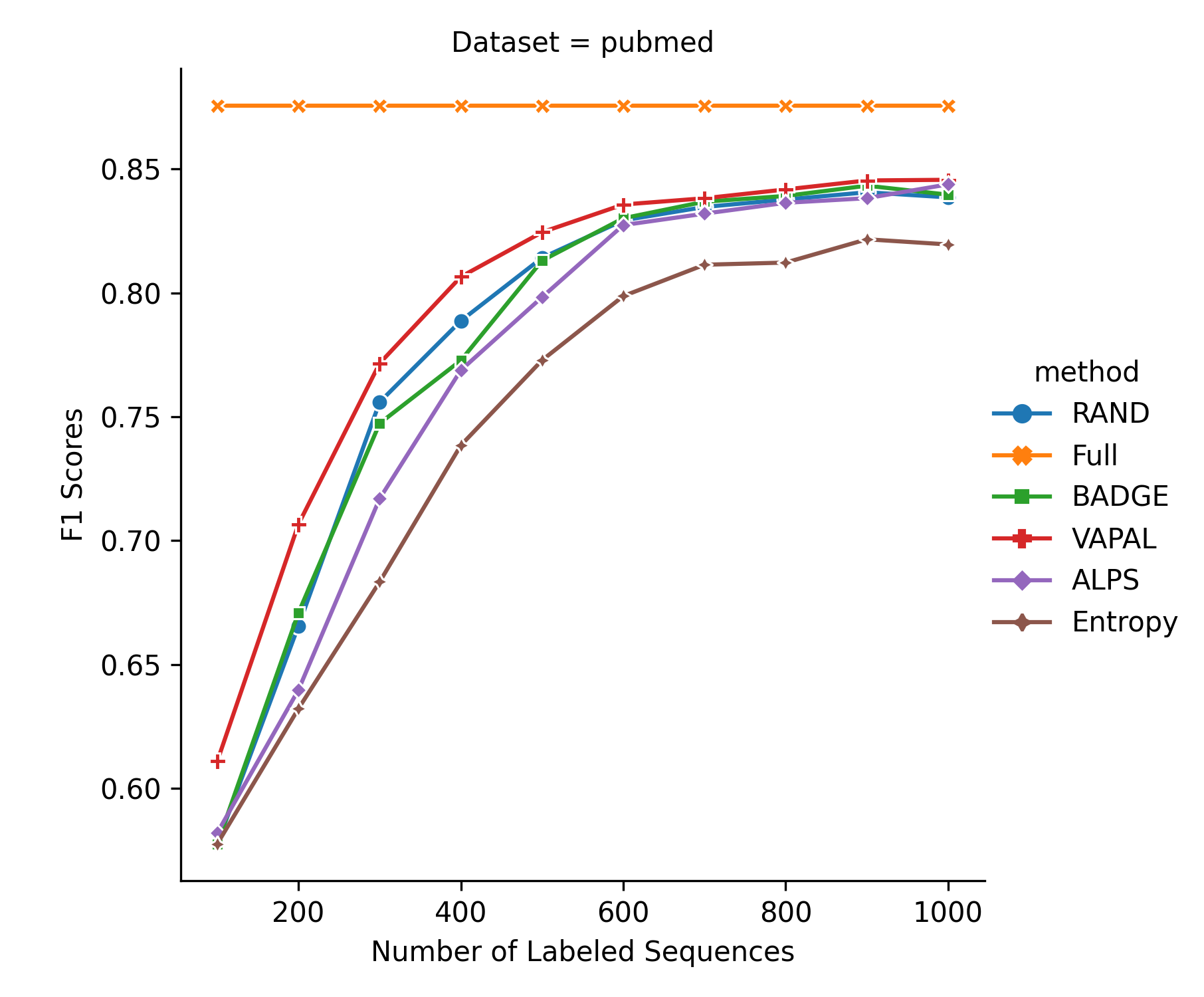}  
   \caption{Dataset: PUBMED}
  \label{fig:whole-pubmed}
\end{subfigure}%
\begin{subfigure}[b]{.5\textwidth}
  % include fourth image
  \centering
  \includegraphics[width=\linewidth]{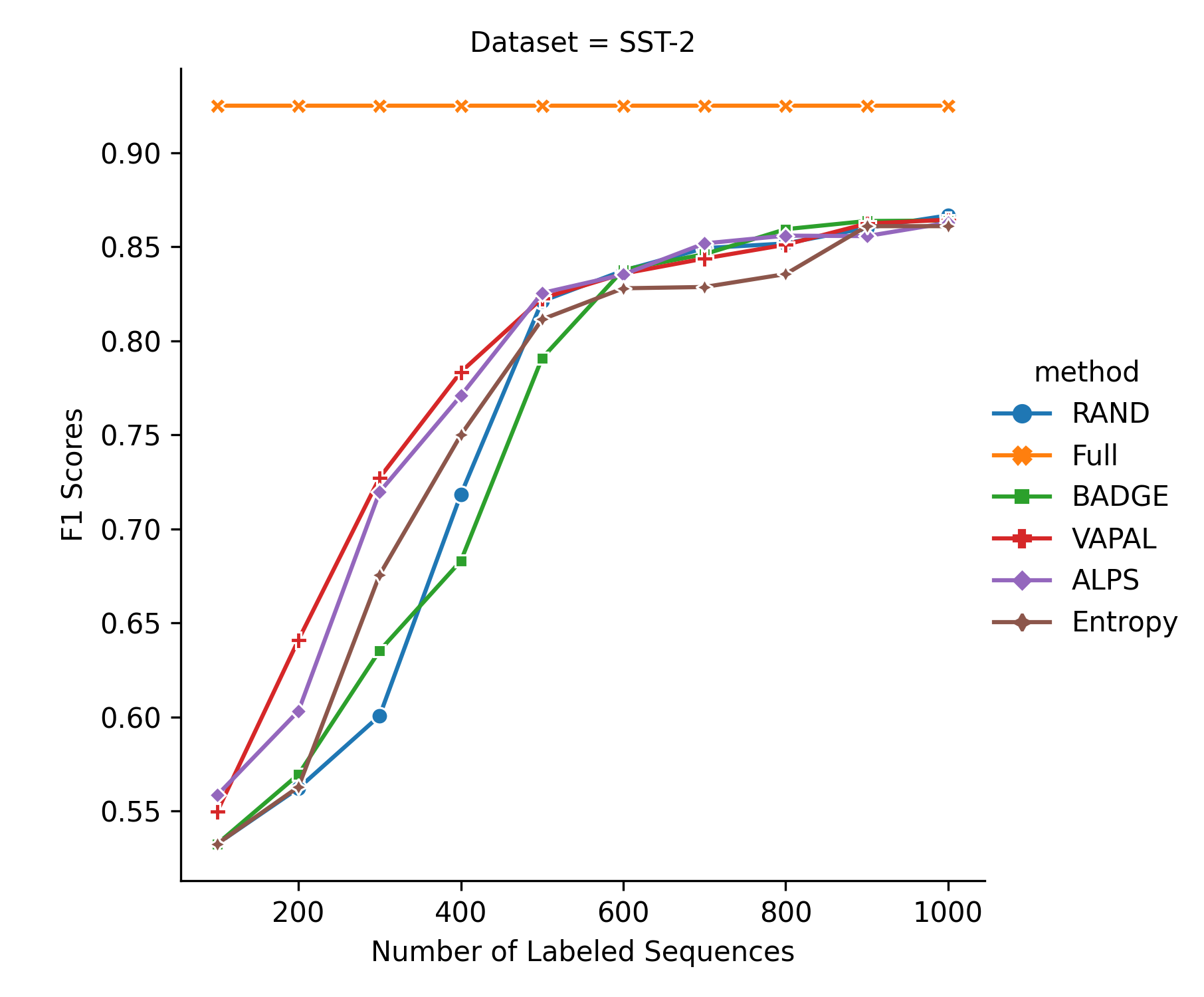}  
  \caption{Dataset: SST-2}
 \label{fig:whole-sst2}
\end{subfigure}

\caption{Average test F1 score of active learning simulation over 10 iterations with 100 sentences per iteration for four data sets. The numbers are the averaged test F1 score of five runs with different random seeds. Overall, VAPAL can select more informative data points for model resulting higher test F1 score especially in PUBMED data set. The fully supervised performance (Full) is also reported.}
\label{fig:whole}
\end{figure*}

The test F1 score across all data sets and baseline acquisition functions is presented in Figure \ref{fig:whole}. VAPAL can perform equally well or even better than other baseline approaches especially on PUBMED, SST-2. Meanwhile on AGNEWS, most methods yield similar results except ENTROPY. The pool performance of Entropy indicates that the uncertainty method based on model confidence score does suffer from the poorly calibrated models. 

Similarly, we can see that BADGE also achieves low performance, even though it uses both uncertainty and diversity. It indicates that the constructed gradient embeddings from the assumed target label do not benefit data sampling. 

On PUBMED and SST-2, VAPAL achieved a larger performance margin than other methods most of the time excluding the early stages of ALPS on SST-2. VAPAL consistently outperforms BADGE, suggesting that our method can be a successful replacement of BADGE. It can also be concluded that ALPS and VAPAL are compatible with each other from Figure.\ref{fig:whole}. 

Unexpectedly, We do not observe the same performance-boosting reported in \cite{Yuan2020Cold-startModeling} between ALPS and BADGE on PUBMED dataset (Figure.\ref{fig:whole-pubmed}). The stability of AL methods against randomness needs to be further explored in the future. The performance differences of ALPS on four datasets, especially the large drop on PUBMED and the second-best performance on SST-2, show the limitation of using perplexity (loss) of pretrained language model as model uncertainty measure. The correlation of perplexity and the difficulty of an example in downstream tasks should be evaluated in future research.

It is interesting to observe that test F1 score across the different methods converges in the end as the sampled data points increase. The performance gap between AL methods and fully supersized is still in need of improvement. This observation highlights the future direction of AL strategy in the early stages.

\subsection{Is $r_{v-adv}$ The Best Choice Applying VAT For AL?}
\label{Appendix:Best vat}
\begin{figure*}[ht]
\begin{subfigure}{.5\textwidth}
  \centering
  % include first image
  \includegraphics[width=\linewidth]{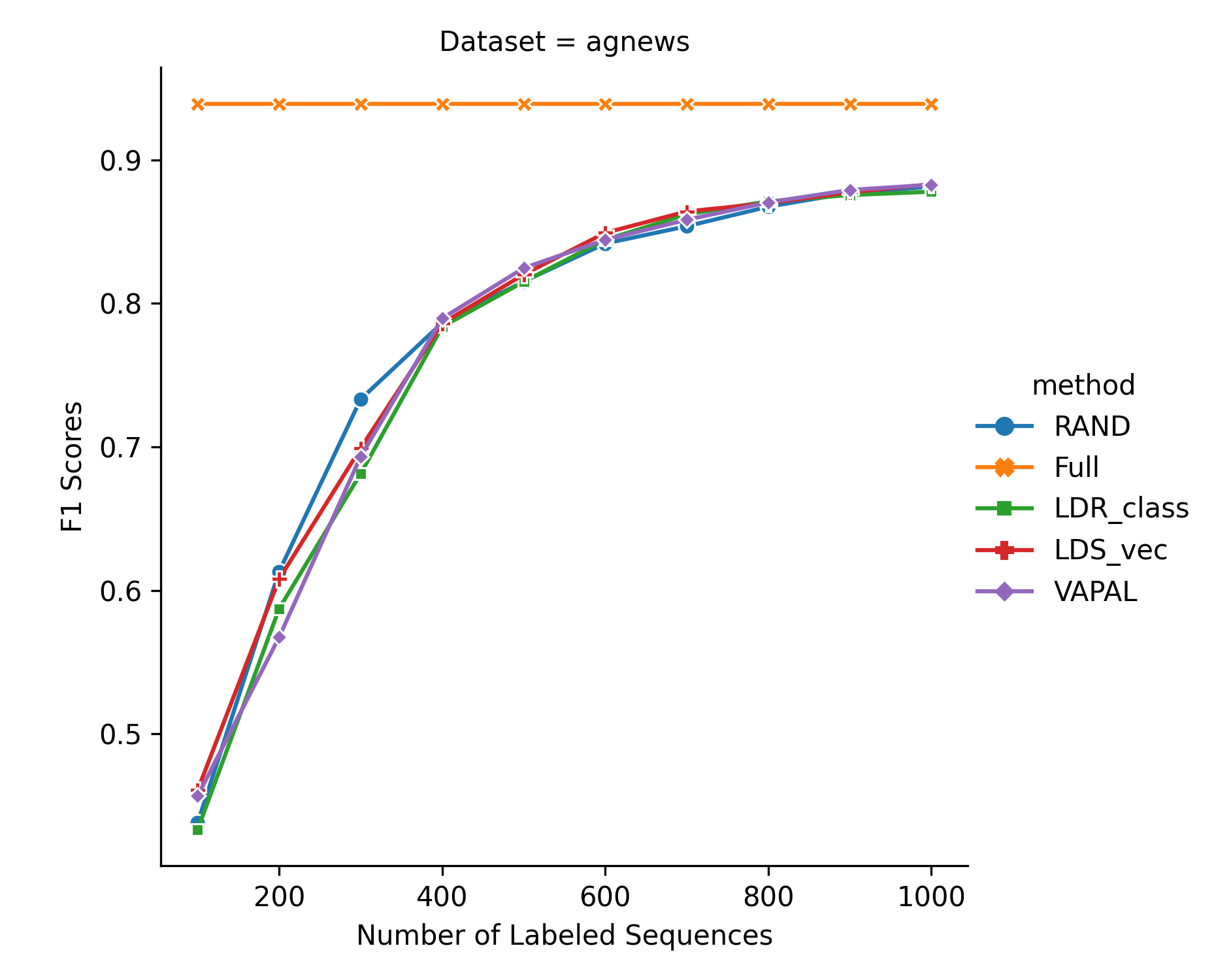}  
  \caption{Dataset: AGNEWS}

  \label{fig:vat-agnews}
\end{subfigure}
\begin{subfigure}{.5\textwidth}
  \centering
  % include second image
  \includegraphics[width=\linewidth]{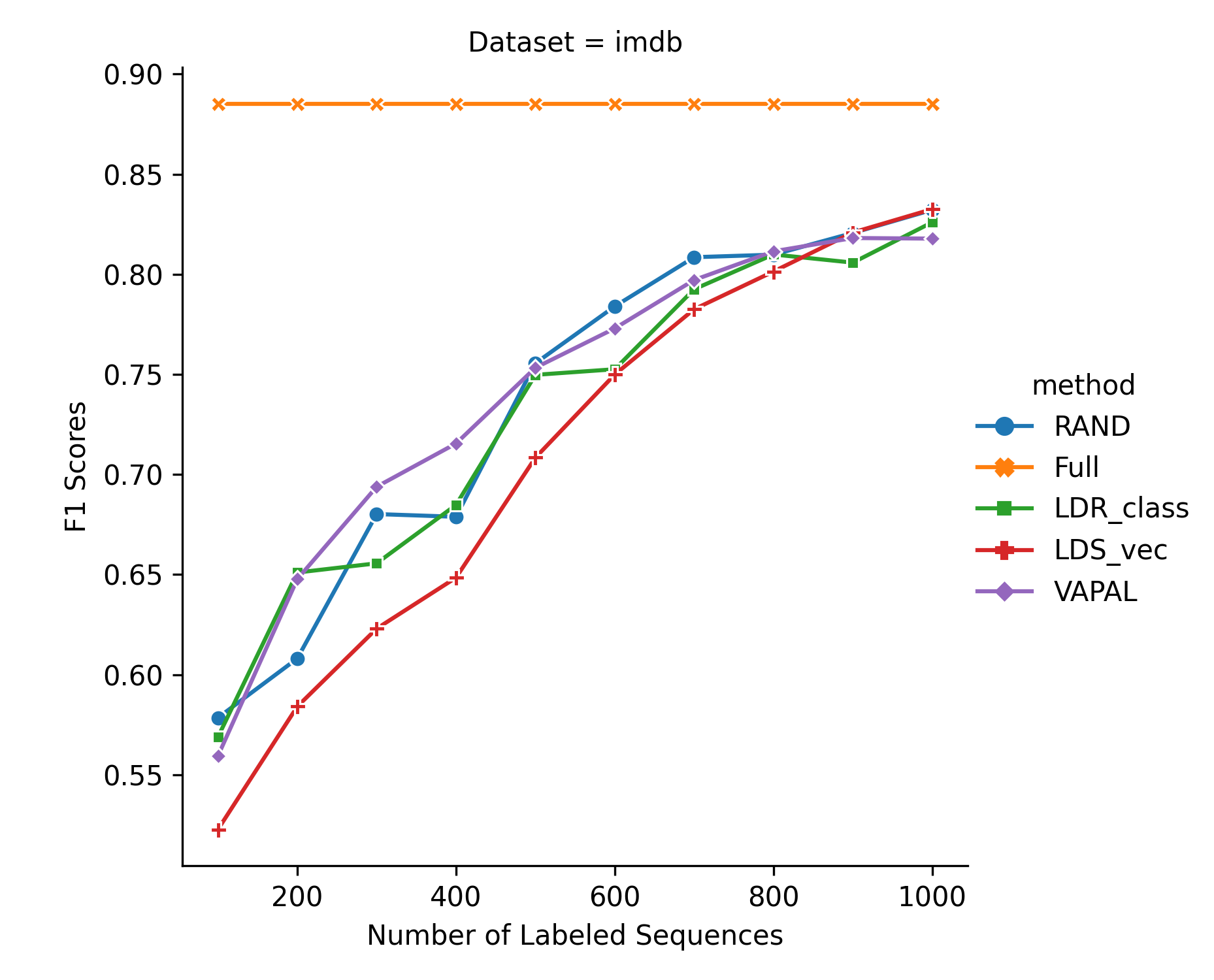}  
  \caption{Dataset: IMDB}
  \label{fig:vat-imdb}
\end{subfigure}
\newline
\begin{subfigure}{.5\textwidth}
  \centering
  % include third image
  \includegraphics[width=\linewidth]{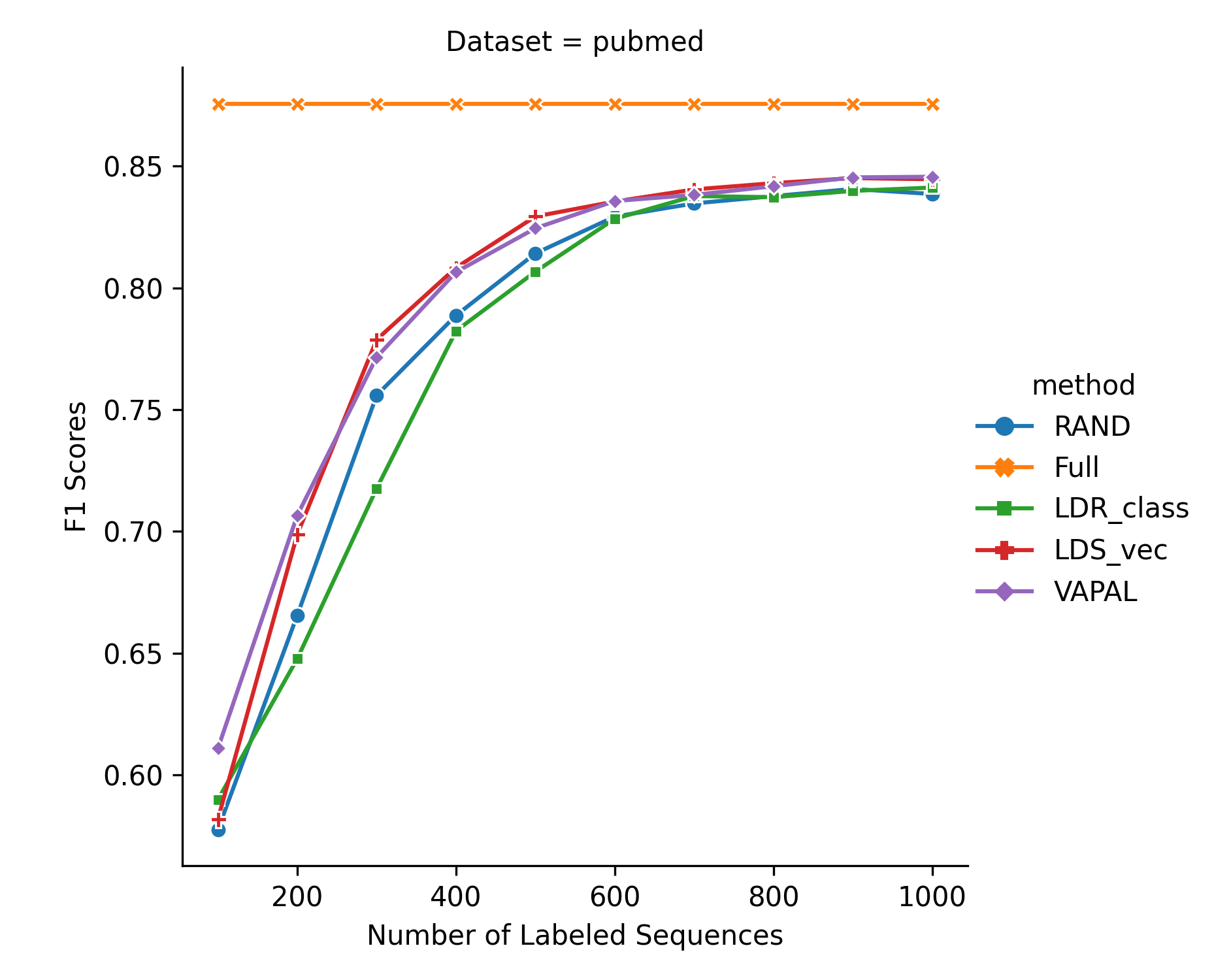}  
    \caption{Dataset: PUBMED}
 \label{fig:vat-pubmed}
\end{subfigure}
\begin{subfigure}{.5\textwidth}
  \centering
  % include fourth image
  \includegraphics[width=\linewidth]{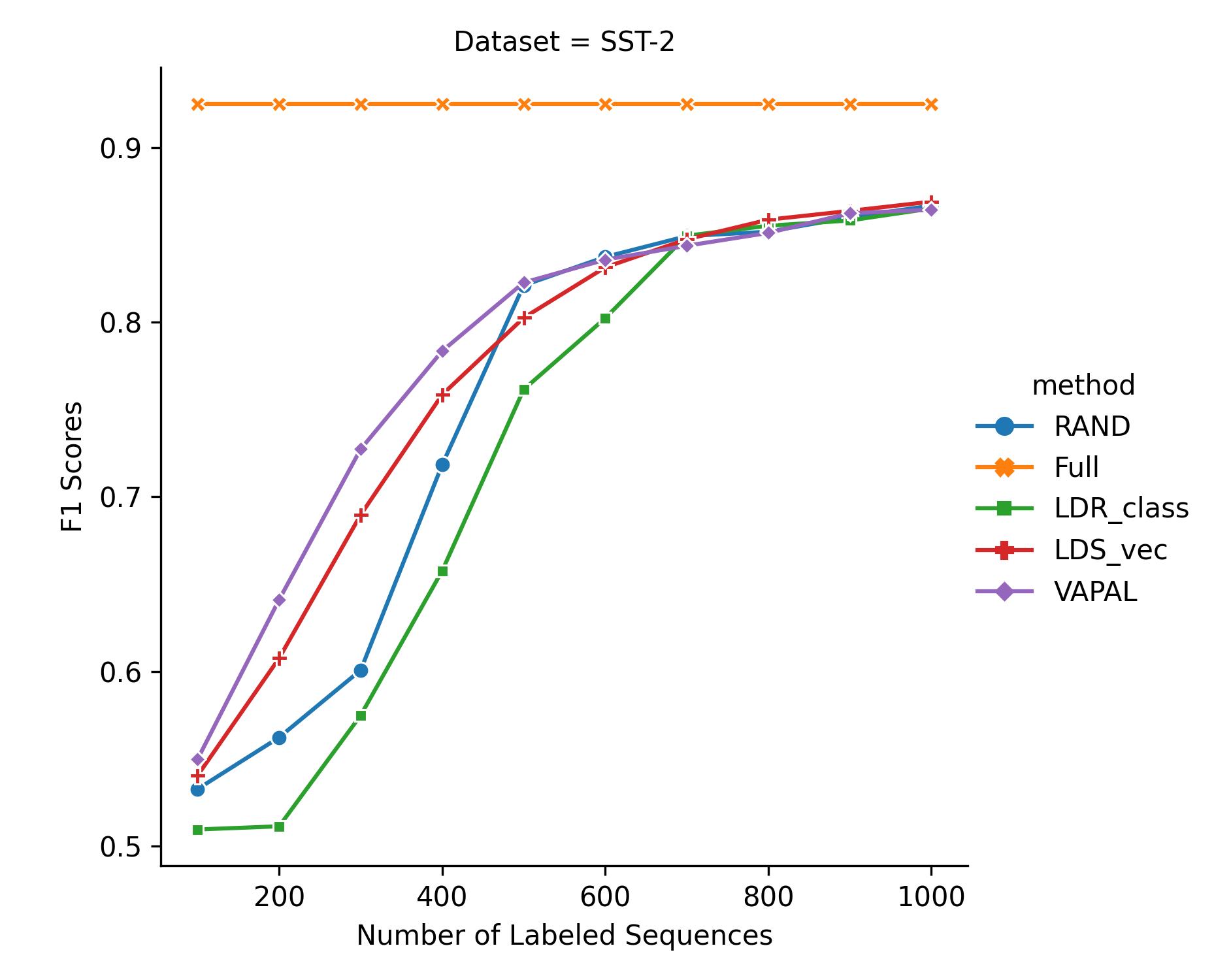}  
 \caption{Dataset: SST-2}
  \label{fig:vat-sst2}
\end{subfigure}
\caption{Evaluating three way of applying VAPAL into active learning. Two vector based data point representation, VAPAL and ${LDS_{vec}}$ are generally outperform the score based ${LDS_{class}}$. }
\label{fig:VAT}
\end{figure*}

Different approaches could be used to effectively introduce the idea of VAT into the uncertainty sampling stage. In this section, we discuss another two different ways compared to VAPAL:

\textbf{$LDR_{class}$} \citet{Yu2020VirtualLearning}
, a method that uses Local Distribution Roughness $LDR(x_i,\theta)=D_{KL}(\mathbf{r}_{v-adv}^i,x_i,\theta)$ as score function which is a measure of model local smoothness given input $x_i$, selects data point which has $LDR_{x_i} \geq PRT$ where $PRT$ is percentile rank threshold ,  groups data points respected to the model predict label $\hat{y}$ and selecting same amount data from these groups regrading their $LDR$ scores. We use $PRT = 90$  for evaluation following the original setting.

\textbf{$LDS_{vec}$} instead of using the average score of KL, We consider the output of KL-divergence loss which is a $C$-way vector as representation for each data point and then apply $k$-MEANS like VAPAL. 

The performance of these two methods and our VAPAL is shown in  Firgure.\ref{fig:VAT}. From the Figure.\ref{fig:VAT}, the $LDS_{vec}$ generally performs better than $LDR_{class}$ in AGNEWS, PUBMED,SST-2 expect IMDB. 

From this observation, we might conclude that the uncertainty-diversity framework, which combines vector representation of model local smoothness and clustering methods, is a better option. Comparing VAPAL and $LDS_{vec}$, the $r_{v-adv}$ is indicated to be the best uncertainty proxy measurement than others.

\subsection{Is Rand Starting Seed Examples Important?}
% 这里对比第一轮使用随机和第一轮直接抽样，VAT，vec对比来说，rand并不会获得明显的提升，score方向会有明显提升。怎样具体的方式做seed selecting 留给未来的方向继续
\begin{figure*}
\begin{subfigure}{.5\textwidth}
  \centering
  % include first image
  \includegraphics[width=\linewidth]{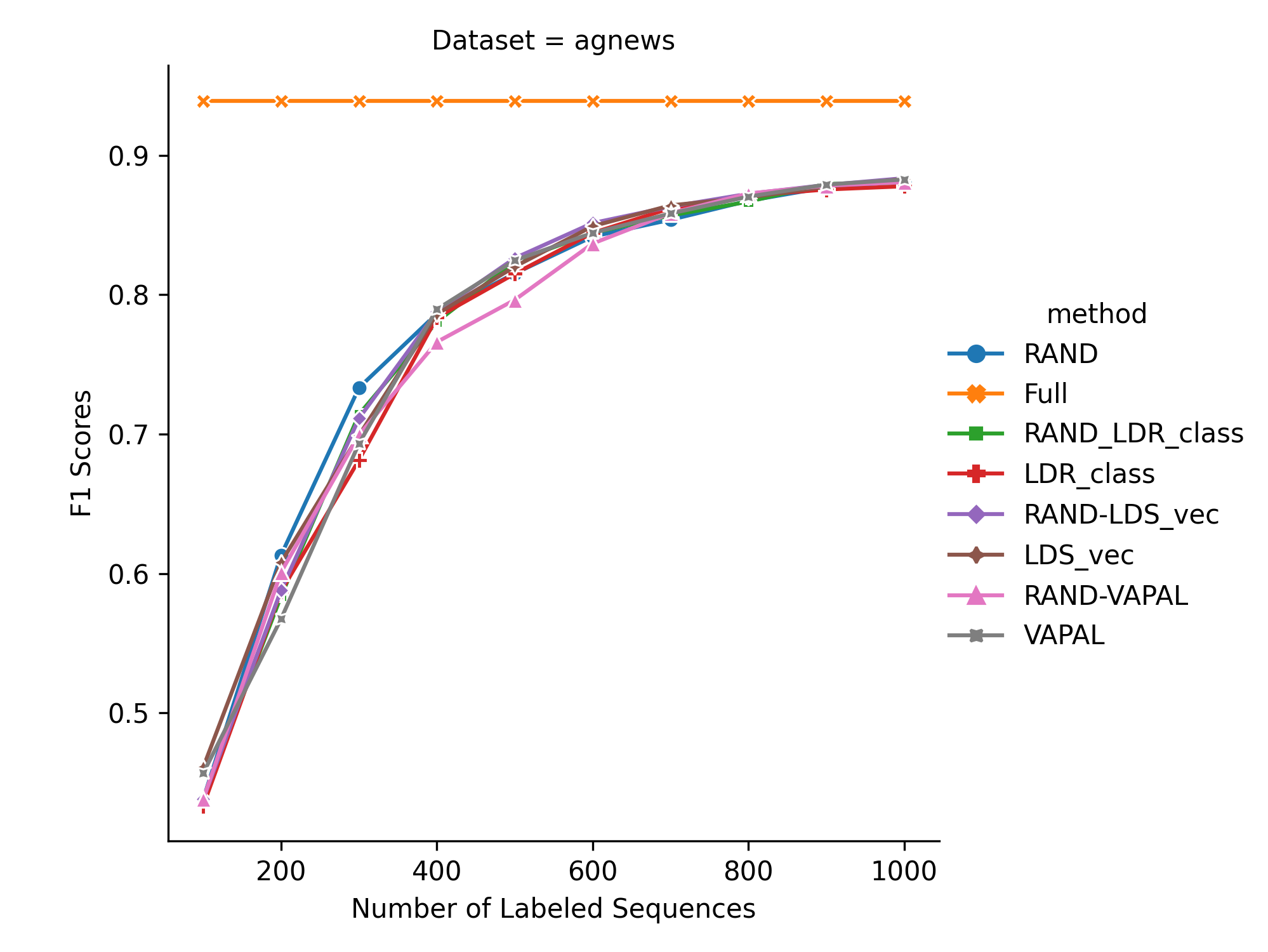}  
  \caption{Dataset: AGNEWS}
  \label{fig:warm-agnews}
\end{subfigure}
\begin{subfigure}{.5\textwidth}
  \centering
  % include second image
  \includegraphics[width=\linewidth]{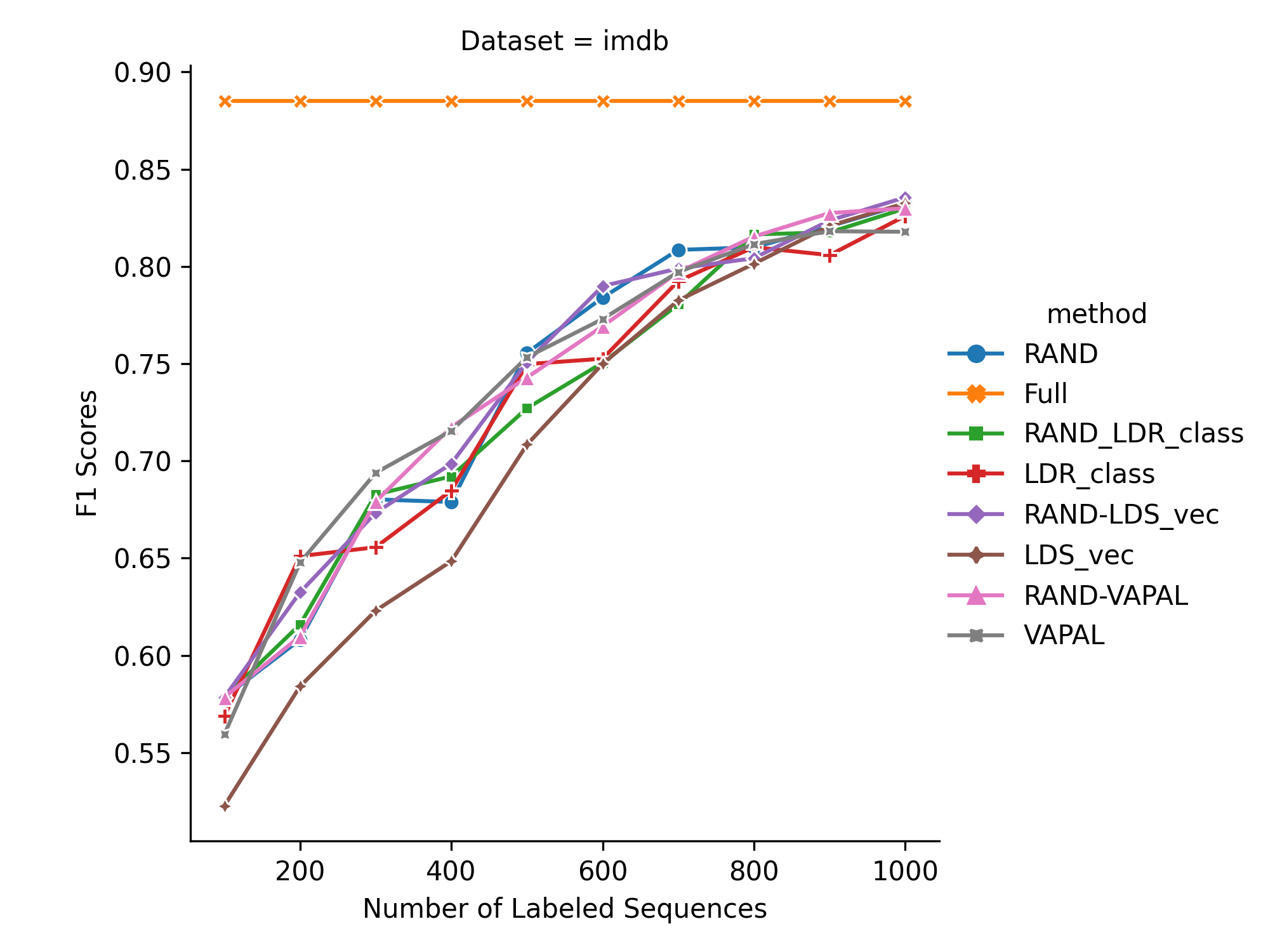}  
  \caption{Dataset: IMDB}
  \label{fig:warm-imdb}
\end{subfigure}

\begin{subfigure}{.5\textwidth}
  \centering
  % include third image
  \includegraphics[width=\linewidth]{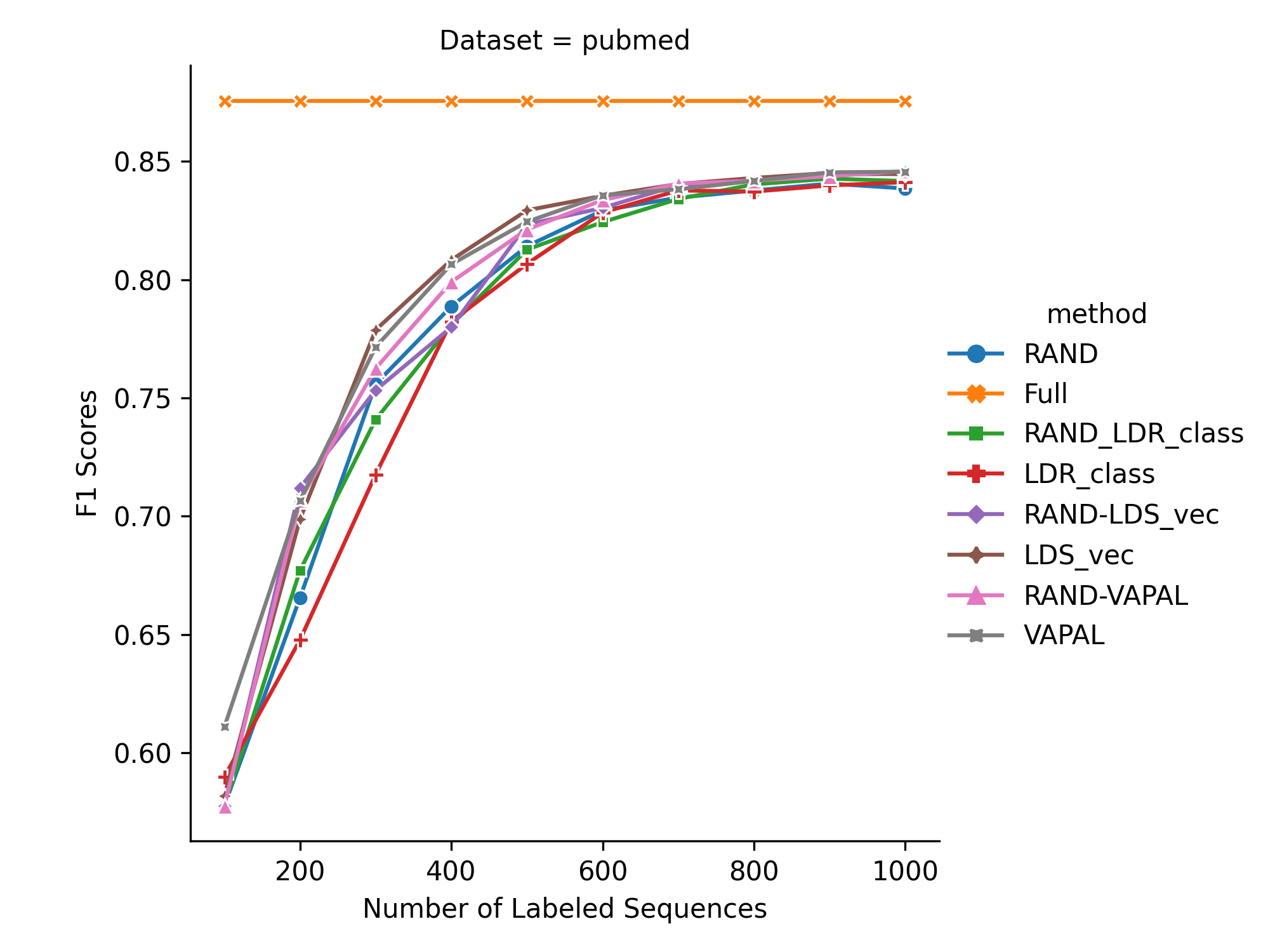}  
  \caption{Dataset: PUBMED}
  \label{fig:warm-pubmed}
\end{subfigure}
\begin{subfigure}{.5\textwidth}
  \centering
  % include fourth image
  \includegraphics[width=\linewidth]{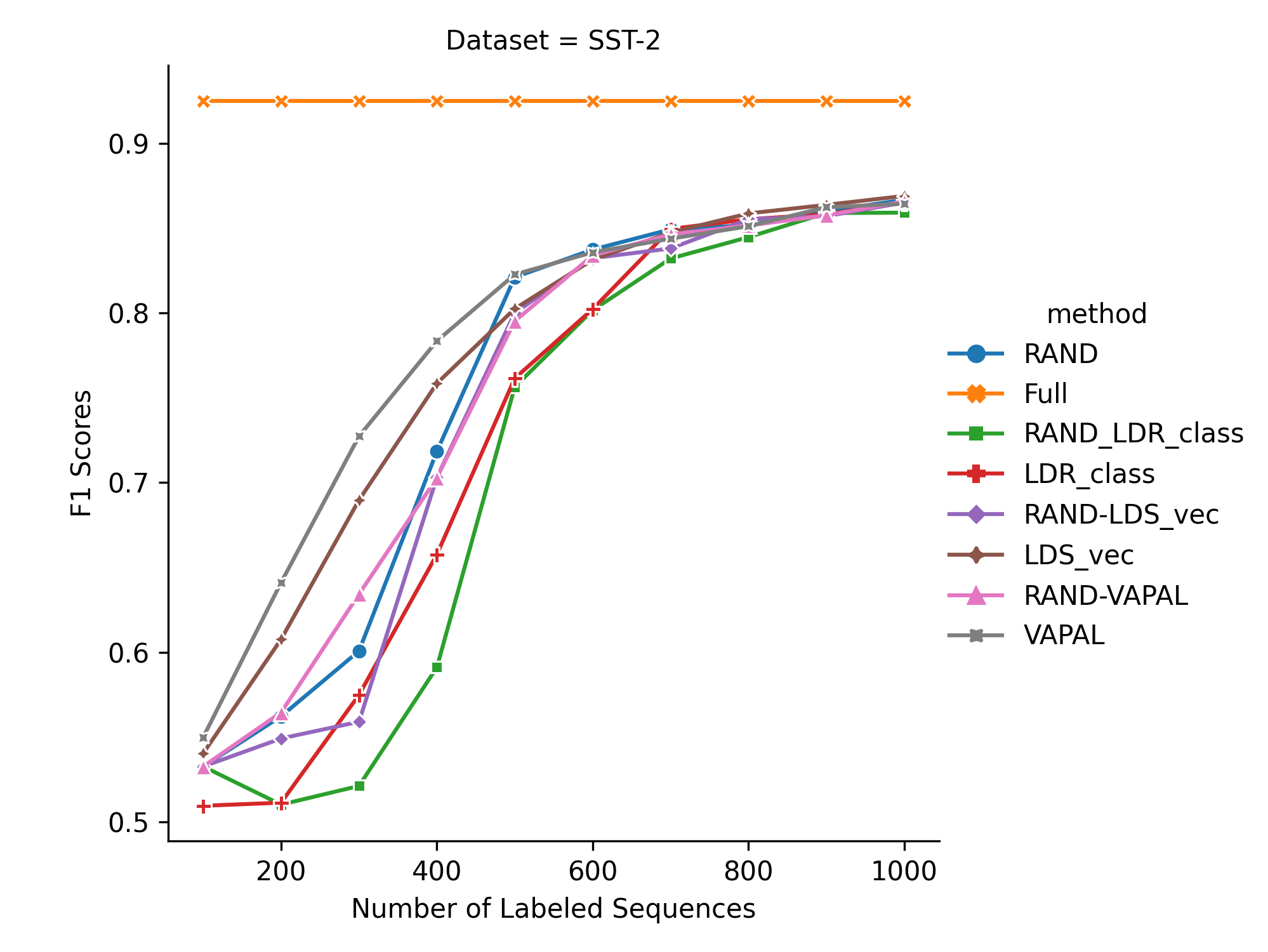}  
  \caption{Dataset: SST-2}
  \label{fig:warm-sst-2}
\end{subfigure}
\caption{Is Rand Starting Seeds Important? RAND-* methods are using the first 100 seeds points from uniform sampling and other methods is directly using proposed sampling strategy }
\label{fig:Rand-Selecting}
\end{figure*}

To answer the question: is random initial starting examples important for virtual adversarial training methods? We experiment to compare two approaches, random selecting from uniform distribution or using the proposed methods with a random initial classification head to get the first hundred seed examples. From Figure.\ref{fig:Rand-Selecting}, VAT methods could select important examples without random initial starting seeds and the VAT methods cloud be suffered from bad initial starting examples. The performance gaps between RANDs and NONE-RANDS under VAPAL method reflects that VAPAL can guide the model training with the help of more informative sampled examples, especially in PUBMERD and SST-2. VAPAL's performance is more stable than others, which indicate the rand starting seed examples might not be important for VAPAL.

\subsection{Can VAPAL apply to GPT2SequenceClassification ?}
\begin{figure*}
\begin{subfigure}{.5\textwidth}
  \centering
  % include first image
  \includegraphics[width=1\linewidth]{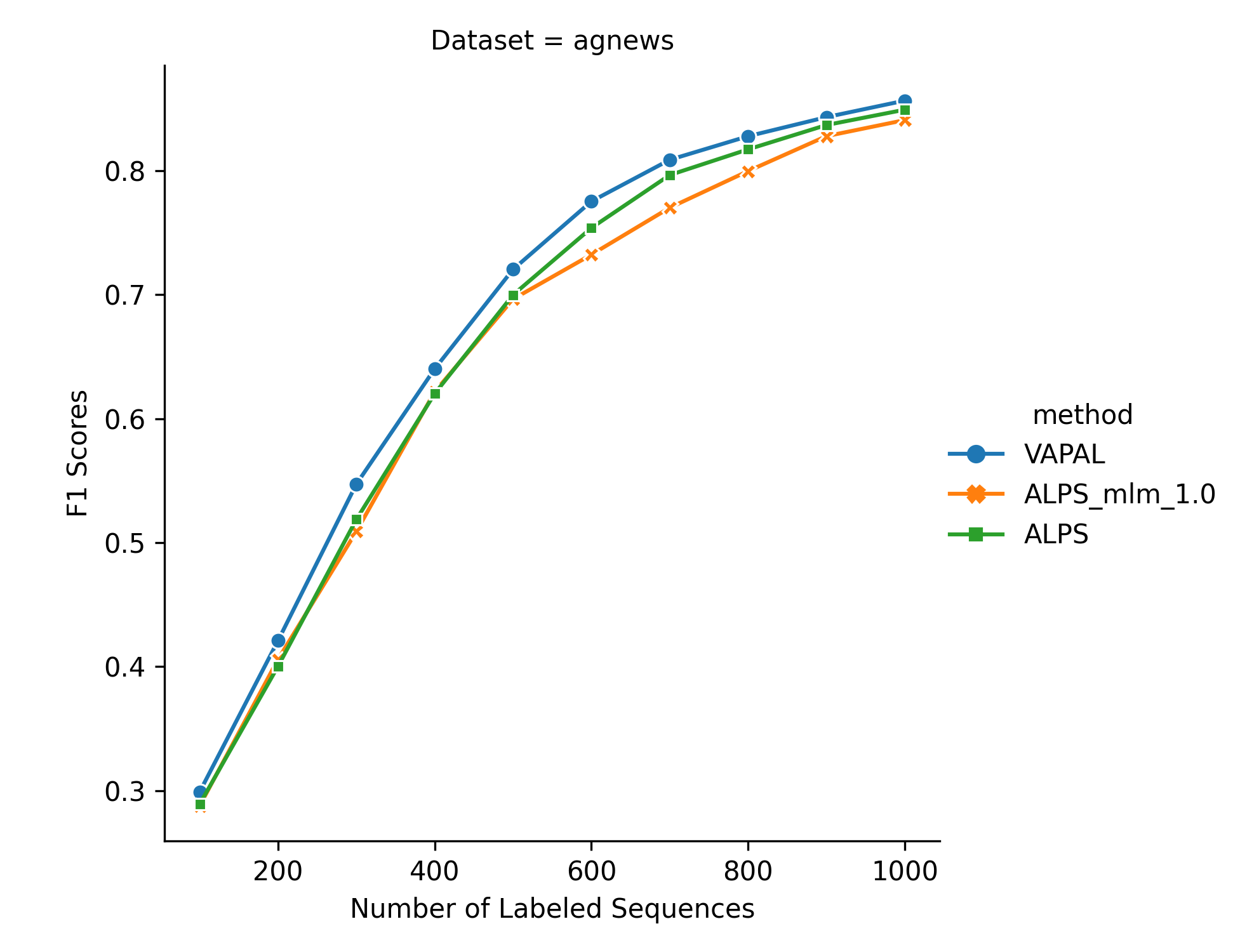}  
  \caption{Dataset: AGNEWS}
  \label{fig:gpt-agnews}
\end{subfigure}
\begin{subfigure}{.5\textwidth}
  \centering
  % include second image
  \includegraphics[width=1\linewidth]{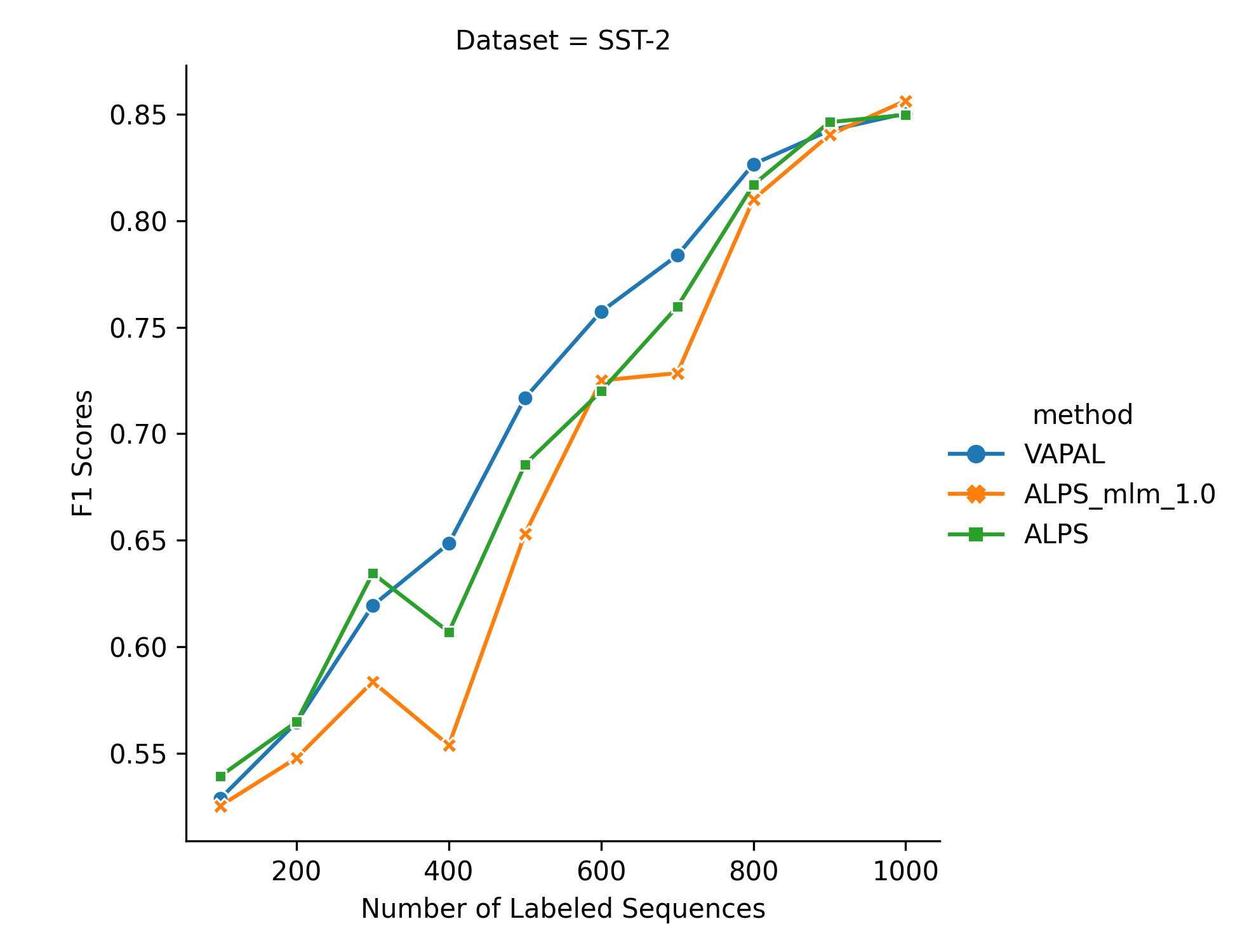}  
  \caption{Dataset: SST-2}
  \label{fig:gpt-SST-2}
\end{subfigure}

\caption{Generalization Ability: The test F1 score results w.r.t the label size in the training with GPT2 as the encoder. The VAPAL is consistently outperform the current state-of-art ALPS and this shows its ability for pre-train causal language model  }
\label{fig:GPT-Generalization}
\end{figure*}

The virtual adversarial perturbation can be applied to any differentiable deep model.  Can it work well with the pre-train causal language model, like GPT2? To answer this question, we evaluate ALPS and VAPAL on the GPT2 \footnote{GPT2 pre-train model is from \url{https://huggingface.co/gpt2}}
\footnote{For classification task, we use sequence classifier from \url{https://huggingface.co/docs/transformers/model_doc/gpt}} on AGNEWS and SST-2, reported in Figure.\ref{fig:GPT-Generalization}. We set two masked probabilities, 15\% and 100\% for ALPS and other settings remain the same. The performance of VAPAL consistently outperforms ALPS in two data sets, especially a large gap in the early stages under SST-2 as shown in Figure.\ref{fig:gpt-SST-2}. 
\section{Conclusion}

 We propose a novel active learning acquisition function called VAPAL. As a proxy of the model's uncertainty, we firstly introduce the virtual adversarial perturbation into active learning for sentence understanding tasks. We use perturbation to represent data points and select the most representative points by clustering following BADGE and ALPS's uncertainty-diversity framework. With the help of virtual adversarial perturbation, we can acquire uncertain examples without the label information. Experiments on various sentence classification tasks and domains demonstrate that VAPAL is another potential option for active learning in sentence understanding tasks. Our empirical results show that further improvement is still required in the future, like Finding more sophisticated local smoothness measurements to encode uncertainty and diversity information for natural language understanding tasks. Another exciting direction is the sensitiveness of active learning methods to random seeds, as we have mentioned.

% Entries for the entire Anthology, followed by custom entries
\bibliography{anthology,custom,references}
\bibliographystyle{acl_natbib}

\appendix
\end{document}